\documentclass[letterpaper]{article} % DO NOT CHANGE THIS
\usepackage{aaai2026}
\usepackage{times}  % DO NOT CHANGE THIS
\usepackage{helvet}  % DO NOT CHANGE THIS
\usepackage{courier}  % DO NOT CHANGE THIS
\usepackage[hyphens]{url}  % DO NOT CHANGE THIS
\usepackage{graphicx} % DO NOT CHANGE THIS
\usepackage{natbib}  % DO NOT CHANGE THIS AND DO NOT ADD ANY OPTIONS TO IT
\usepackage{caption} % DO NOT CHANGE THIS AND DO NOT ADD ANY OPTIONS TO IT
\usepackage{booktabs}
\usepackage{algorithm}
\usepackage{algorithmic}
\usepackage{amssymb}
\usepackage{newfloat}
\usepackage{listings}
\DeclareCaptionStyle{ruled}{labelfont=normalfont,labelsep=colon,strut=off} % DO NOT CHANGE THIS
\floatstyle{ruled}
\newfloat{listing}{tb}{lst}{}
\floatname{listing}{Listing}
\usepackage[utf8]{inputenc}
\usepackage{amsmath}

\title{PRNet: Original Information Is All You Have}
\author{
    PeiHuang Zheng,
    Yunlong Zhao, 
    Zheng Cui, 
    Yang Li*
}
\affiliations {
    \textsuperscript{\rm }Nanjing University of Aeronautics and Astronautics \\
    \{bluce\_8185,zhaoyunlong,cuizheng,liyangnuaa\}@nuaa.edu.cn
}

\usepackage{bibentry}
\begin{document}

\maketitle

\begin{abstract}
Small object detection in aerial images suffers from severe information degradation during feature extraction due to limited pixel representations, where shallow spatial details fail to align effectively with semantic information, leading to frequent misses and false positives. Existing FPN-based methods attempt to mitigate these losses through post-processing enhancements, but the reconstructed details often deviate from the original image information, impeding their fusion with semantic content. To address this limitation, we propose PRNet, a real-time detection framework that prioritizes the preservation and efficient utilization of primitive shallow spatial features to enhance small object representations. PRNet achieves this via two modules:the Progressive Refinement Neck (PRN) for spatial-semantic alignment through backbone reuse and iterative refinement, and the Enhanced SliceSamp (ESSamp) for preserving shallow information during downsampling via optimized rearrangement and convolution. Extensive experiments on the VisDrone, AI-TOD, and UAVDT datasets demonstrate that PRNet outperforms state-of-the-art methods under comparable computational constraints, achieving superior accuracy-efficiency trade-offs.

\end{abstract}

\begin{links}
    \link{Code}{https://github.com/hhao659/PRNet}
\end{links}
\section{Introduction}

Small object detection in aerial imagery has become increasingly important in remote sensing and computer vision, enabling critical applications such as traffic monitoring (\citeyear{kaleem2018amateur}), rescue operations (\citeyear{ren2022improved}), and precision agriculture (\citeyear{roy2022real}). These applications often require real-time inference on resource-limited edge devices while maintaining high accuracy for objects that occupy very few pixels and are challenging to discern.

\begin{figure}[t]
\centering
\includegraphics[width=0.48\textwidth]{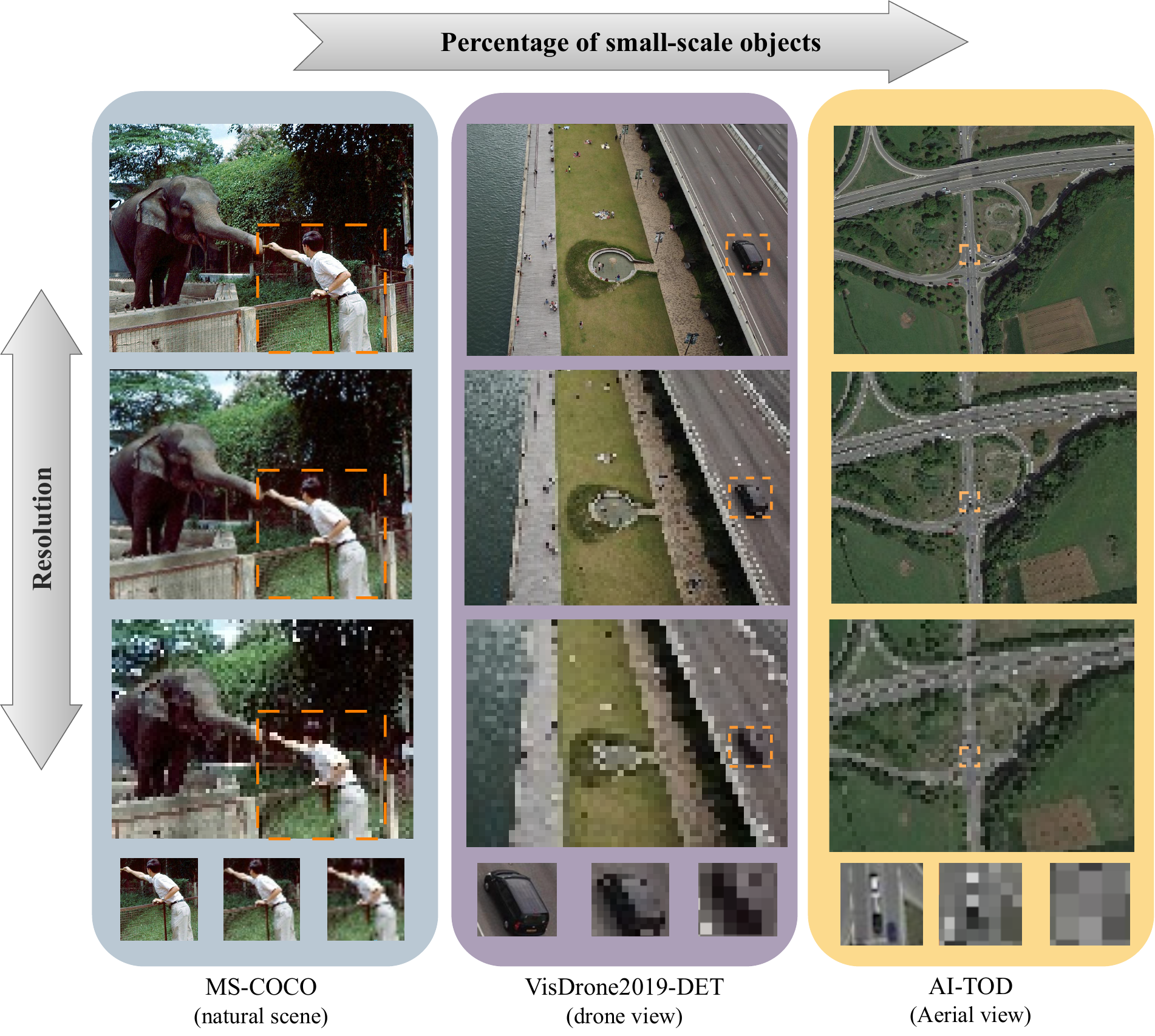} % Reduce the figure size so that it is slightly narrower than the column.
\caption{\textbf{Comparative Analysis of Resolution Degradation on Object Visibility Across Datasets.} Comparison of object visibility degradation across MS-COCO, VisDrone, and AI-TOD at original, 160×160, and 80×80 resolutions (top to bottom). Small objects exhibit greater impact from losses in edges, textures, and shapes during degradation.}
\label{fig:resolution_comparison}
\end{figure}

Detecting small objects in aerial images is fundamentally difficult due to their extremely limited pixel representation and complex, cluttered backgrounds. Unlike natural scene detection, where objects typically occupy substantial portions of the image, aerial objects are exceptionally small—often under 32×32 pixels and occupying merely 0.1\% to 1\% of the total image area (\citeyear{lin2014microsoft}). As demonstrated in Figure~\ref{fig:resolution_comparison}, when image resolution decreases (from the original resolution to 160×160 and 80×80), small objects suffer catastrophic information loss compared to larger objects. This phenomenon mirrors the information degradation that occurs during the model's forward propagation (\citeyear{deng2021extended,bian2025refined}), where the loss of shallow spatial details leads to semantic mismatches and, consequently, increased rates of false positives and missed detections (\citeyear{xiao2025fbrt,hua2025survey}).

Contemporary object detection models comprise a backbone, neck, and head, with the neck—commonly known as the Feature Pyramid Network (FPN,\citeyear{lin2017feature})—serving as the primary architecture for multi-scale feature aggregation (\citeyear{bouraya2021deep}). FPN have become a foundational framework for multi-scale object detection due to their ability to aggregate features across different resolutions. Nevertheless, traditional FPNs are suboptimal for small object detection in aerial imagery. This limitation stems from their reliance on fusing features that have already undergone multiple convolutional and sampling operations, resulting in substantial loss of high-resolution spatial details critical for identifying small objects.

% Recent FPN variants (\citeyear{shi2025hs,xiao2025fbrt, wang2025efcnet}) primarily focus on enhanced feature fusion or additional refinement modules. However, these approaches fail to effectively mitigate the progressive information degradation that accumulates during feature extraction, particularly in its early stages. Specifically, FPN-based methods face three critical challenges:
% (1) Accumulated information degradation: Progressive sampling operations and convolutions cause irreversible loss of fine-grained details that cannot be fully restored through upsampling or conventional feature fusion.
% (2) Underutilization of shallow features: High-resolution shallow features (e.g., at the P$_2$ level) are typically used only once, leading to a permanent loss of spatial information crucial for small object discrimination.
% (3) Feature misalignment: The difficulty of aligning shallow spatial features with deeper semantic representations often results in mismatches, thereby reducing detection effectiveness.
Recent FPN variants (\citeyear{shi2025hs,xiao2025fbrt, wang2025efcnet}) primarily focus on enhanced feature fusion or additional refinement modules. However, these approaches do not adequately mitigate the progressive information degradation that accumulates during feature extraction, particularly in its early stages, leading to an irreversible loss of fine-grained details that cannot be fully restored through upsampling or conventional feature fusion. Specifically, FPN-based methods face two critical challenges: (1) Underutilization of shallow features: High-resolution shallow features (e.g., at the P$_2$ level) are typically used only once, leading to a permanent loss of spatial information crucial for small object discrimination. (2) Feature misalignment: Difficulty in fully integrating shallow spatial features with deeper semantic representations for small targets, thereby resulting in feature mismatches that reduce detection effectiveness.

To address these challenges, preserving and effectively utilizing high-resolution information from initial processing is critical. We propose PRNet, a novel framework tailored for aerial small object detection. First, to maximize the use of preserved shallow features, PRNet introduces the Progressive Refinement Neck (PRN), which iteratively refines high-resolution features through multi-stage backbone reuse, ensuring robust small object representation while alleviating feature misalignment. PRN is flexible and can be integrated into various detection frameworks. Second, to mitigate detail loss during downsampling, PRNet employs the Enhanced SliceSamp (ESSamp) module, which optimizes spatial rearrangement and enhances depthwise convolution for superior feature preservation.

Our contributions can be summarized as follows:

\begin{itemize}

    \item We reveal key limitations of existing FPN-like methods, specifically information degradation and feature misalignment, which make them unsuitable for aerial image datasets.
    
    \item We design a novel neck architecture, PRN, which achieves efficient high-resolution detail retention through multi-stage feature reuse and progressive fusion. In addition, we develop an enhanced downsampling module, ESSamp, to improve the preservation of shallow spatial information.
    
    \item Experimental results show that our proposed PRNet significantly surpasses state-of-the-art methods, achieving superior detection accuracy while maintaining competitive efficiency.
    
\end{itemize}

\section{Related Works}

\subsection{Small Object Detection}
Object detection in aerial images is a representative small object detection task and has consistently posed a challenge. FFCA-YOLO (\citeyear{10423050}) proposes a context-aware detection framework for remote sensing images, enhancing the model’s ability to perceive semantic context. SFFEF-YOLO (\citeyear{bai2025sffef}) introduces a fine-grained feature extraction module to replace standard convolutions, aiming to reduce information loss during the sampling process. FBRT-YOLO (\citeyear{xiao2025fbrt}) incorporates a Feature Complementary APping Module and a Multi-Kernel Perception Unit to improve semantic alignment and multi-scale object perception, achieving a better trade-off between detection accuracy and efficiency. Nevertheless, high-precision real-time detection of small objects remains a challenging task.

\subsection{Feature Pyramid Networks}
Feature Pyramid Networks (FPNs) are the dominant architecture for multi-scale detection. The original FPN design integrates deep semantic features with shallow spatial features via top-down pathways and lateral connections. Subsequent improvements, such as PANet (\citeyear{liu2018path}), which adds bottom-up pathways, and BiFPN (\citeyear{tan2020efficientdet}), which employs weighted bidirectional fusion, enhance integration efficiency. For small objects, DSP-YOLO (\citeyear{zhang2024dsp}) introduces a lightweight, detail-sensitive DsPAN, while E-FPN (\citeyear{li2025fpn}) enhances semantic and fine-grained details bidirectionally. However, the issue of detail loss cannot be alleviated. Unlike these methods, which refine fusion post-extraction, our PRN iteratively reuses backbone features and applies progressive fusion to preserve high-resolution details, minimizing information loss for small object detection.

\begin{figure*}[t]
\centering
\includegraphics[width=0.9\textwidth]{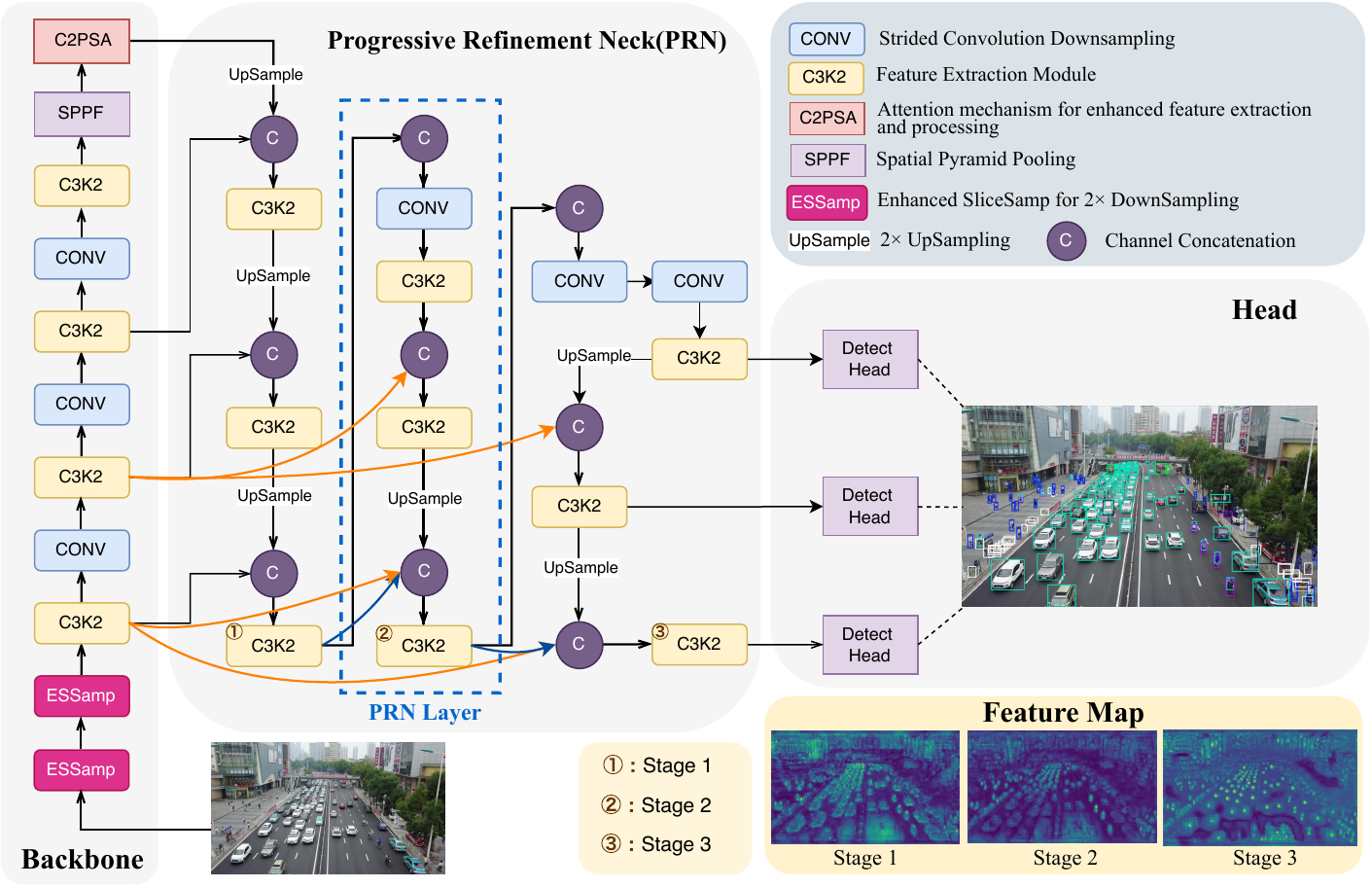} % Reduce the figure size so that it is slightly narrower than the column.
\caption{\textbf{Architecture of Progressive Refinement Network.} Using YOLO11 as the baseline model, we replace PAN-FPN with our proposed PRN and replace traditional stride convolution downsampling with the proposed ESSamp in the first two layers of the network. The bottom left shows comparisons of feature APs at different stages, demonstrating that the feature quality improves as the number of stages increases.}
\label{fig:prnet_architecture}
\end{figure*}

\subsection{Feature-Preserving Downsampling}
The ability of downsampling methods to retain critical information plays a vital role in overall model performance, especially for tiny objects. Content-Adaptive Downsampling (\citeyear{hesse2023content}) attempts to preserve key regions during subsampling, but relies heavily on accurate importance masks, which are difficult to generate for small objects in complex aerial scenes. SliceSamp (\citeyear{he2023slicesamp}) leverages spatial slicing and depthwise separable convolutions to improve computational efficiency while better preserving feature information. DiffStride (\citeyear{rafif2023hybrid}) introduces learnable stride parameters to adaptively control resolution loss, but its increased model complexity limits deployment in resource-constrained environments. While these methods have made notable progress, effectively preserving detailed features during downsampling remains a significant challenge.

\section{Methodology}
In this section, we present a detection framework built upon YOLO11 (\citeyear{khanam2024yolov11}), named PRNet. PRNet's design follows a cohesive pipeline: first, ESSamp optimizes downsampling in the backbone to preserve shallow spatial details; second, PRN iteratively refines these features in the neck through backbone reuse and progressive fusion. Figure~\ref{fig:prnet_architecture} illustrates the overall architecture where PRN replaces traditional PAN-FPN and ESSamp replaces the first two stride convolutions in the backbone. ESSamp complements PRN by ensuring high-quality shallow inputs for reuse.

\subsection{Progressive Refinement Neck}
\begin{figure}[t]
\centering
\includegraphics[width=0.48\textwidth]{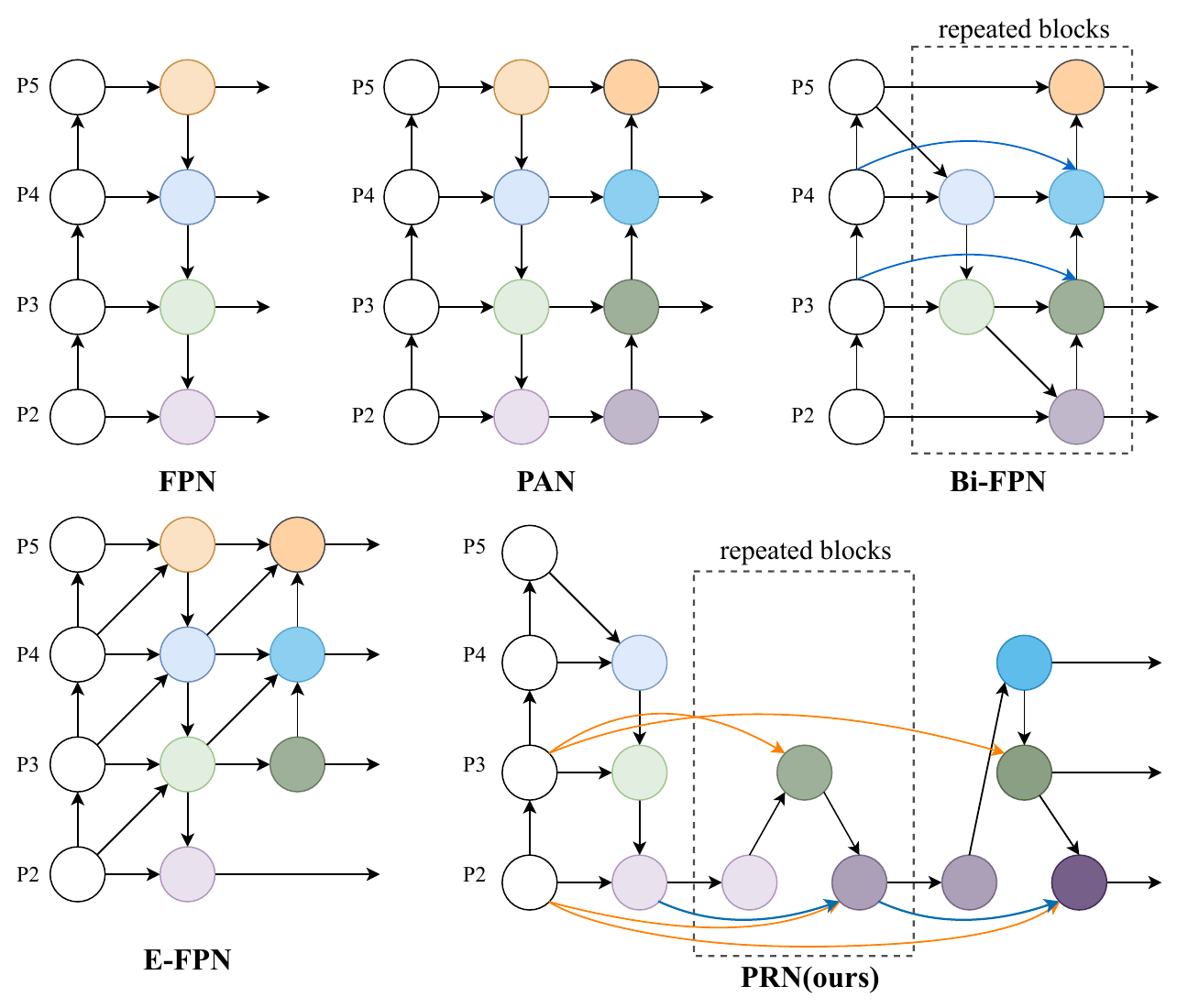} % Reduce the figure size so that it is slightly narrower than the column.
\caption{\textbf{Comparison of PRN and Traditional FPN Architectures.} PRN enables backbone feature reuse (orange lines) and progressive fusion (blue lines) for iterative high-resolution feature refinement.} 
\label{fig:prn_comparison}
\end{figure}

Traditional FPN-based methods suffer from insufficient utilization of high-resolution backbone features. As illustrated in Figure~\ref{fig:prn_comparison}, shallow features containing critical spatial details are typically used only once during fusion. This single-use pattern limits the exploitation of preserved high-resolution information, potentially leading to suboptimal feature representations for small object detection. To address this limitation, we propose the Progressive Refinement Neck (PRN). This module maximizes the retention of original high-resolution details through multi-stage backbone feature reuse, enabling iterative refinement to fully exploit detailed information for enhanced small object representation. The detailed structure of PRN is shown in Figure~\ref{fig:prnet_architecture}, which utilizes a backbone feature reuse mechanism and progressive fusion strategy. The implementation of this module is detailed below.

\noindent\textbf{Initial Feature Fusion.} PRN begins with standard top-down fusion as in PAN-FPN to establish initial spatial-semantic integration:

\begin{equation}
P_i^{td} = {Conv}\{{Concat}({Resize}(P_{i+1}), P_i^{in})\}, \quad i \in \{2,3,4\}
\end{equation}

where Resize denotes upsampling or downsampling operations for resolution matching, and Conv represents convolution operations for feature processing (using 3×3 kernels). This initial fusion maintains compatibility with the baseline YOLO11 while establishing a foundation for subsequent refinement stages.

\noindent\textbf{Backbone Feature Reuse Mechanism.} We noticed the information value of backbone features: The shallow features $P_{2}^{in}$ and $P_{3}^{in}$ from the backbone network contain relatively unprocessed original detail information, which is discarded after single use in traditional FPN, causing information waste. To this end, PRN compensates for information dilution in traditional fusion through multi-stage backbone feature reuse. Specifically, PRN downsamples the top-down fused $P_{2}^{td}$, then concatenates it with the unused backbone feature $P_{3}^{in}$, reintroducing mid-level details; subsequently, the result is upsampled and concatenated with the similarly pristine backbone feature $P_{2}^{in}$, maximizing the utilization of high-resolution original details. As shown in Equations below:
\begin{equation}
P_{3}^{td^1} = Conv\{Concat(Resize(P_{2}^{td}), P_{3}^{in})\}
\end{equation}
\begin{equation}
P_{2}^{refine^1} = Conv\{Concat(Resize(P_{3}^{td^1}), P_{2}^{td}, P_{2}^{in})\}
\end{equation}

After initial feature fusion and subsequent processing, the fused features contain sufficient semantic information. Excessive up-and-down sampling operations would dilute the scarce spatial information critical for small object detection. Therefore, we employ a single downsample-upsample strategy to maximize spatial detail preservation while maintaining computational efficiency.

\noindent\textbf{Progressive Fusion Strategy.} PRN integrates refined features from earlier stages (e.g., $P_{2}^{td}$) into subsequent computations, thereby guiding the refinement process and preventing indiscriminate feature fusion by introducing contextual constraints. This progressive design enables high-resolution features to be iteratively enhanced across multiple stages while remaining consistent with deeper semantic representations. As illustrated by the blue connections in Figure~\ref{fig:prn_comparison}, these progressive links form a gradually optimizing closed loop. To ensure efficient refinement, the progressive fusion process is structured into repeated blocks, where each block reuses backbone features and performs a downsampling–upsampling cycle. This design guarantees that each refinement stage receives original feature inputs from the backbone, mitigating recursive information decay and allowing semantic representations to be progressively enriched while high-resolution details are preserved.

\noindent\textbf{Output Generation.} To generate output features suitable for three detection scales, PRN performs structured processing on the final refined features, as shown in Equations (4)-(6):
\begin{equation}
P_{4}^{out} = Conv\{Resize(P_{2}^{refine^i})\}
\end{equation}
\begin{equation}
P_{3}^{out} = Conv\{Concat(Resize(P_{4}^{out}) , P_{3}^{in})\}
\end{equation}
\begin{equation}
P_{2}^{out} = Conv\{Concat(Resize(P_{3}^{out}) , P_{2}^{refine^i} , P_{2}^{in})\}
\end{equation}

\begin{figure}[t]
\centering
\includegraphics[width=0.48\textwidth]{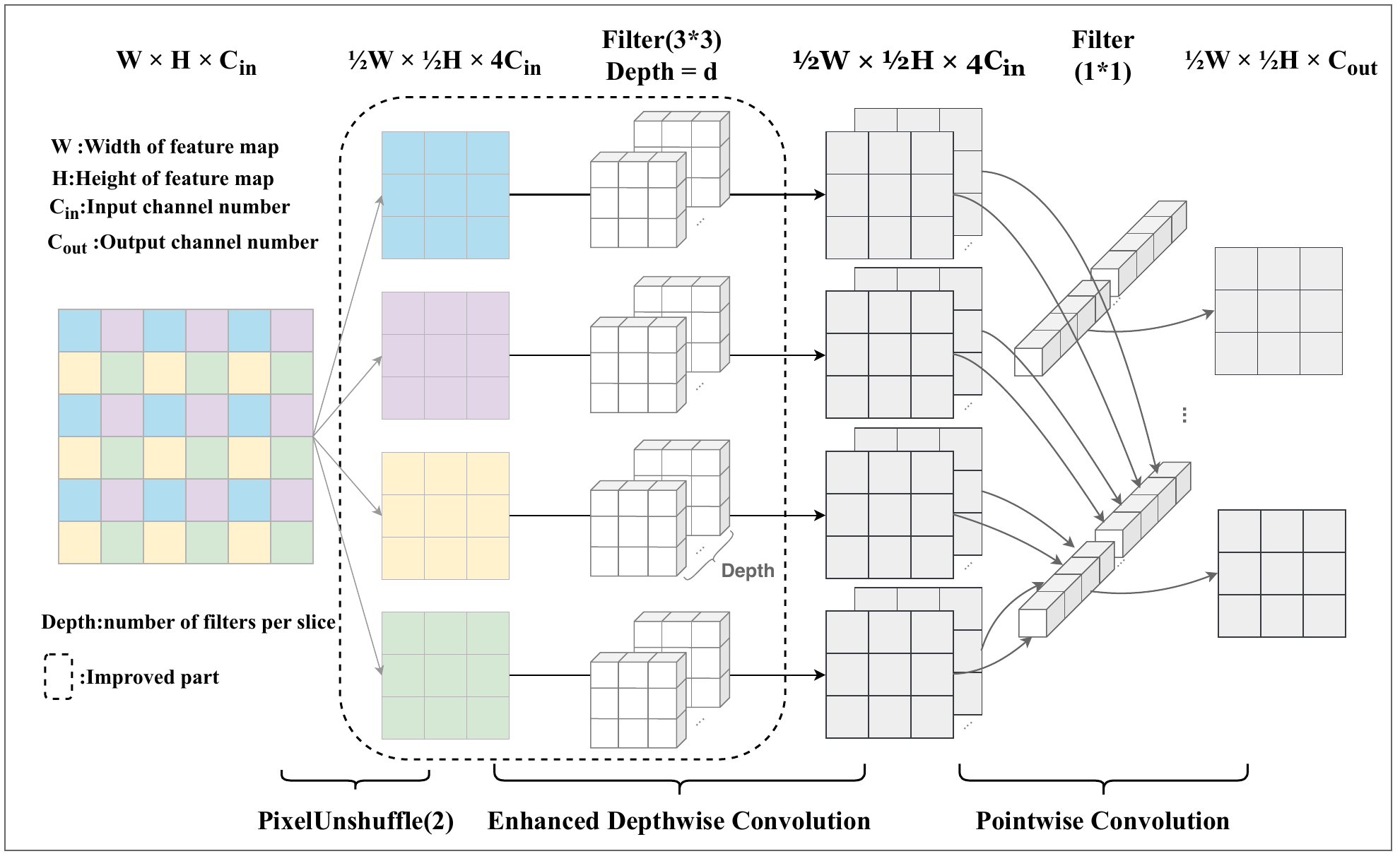} % Reduce the figure size so that it is slightly narrower than the column.
\caption{\textbf{ESSamp Module Structure.} Utilizes PixelUnShuffle for efficient spatial rearrangement and augmented depthwise convolution (depth multiplier d=2) to enhance feature expressiveness, preserving fine-grained details for small object detection.}
\label{fig:essamp_structure}
\end{figure}

\subsection{Enhanced SliceSamp}

While PRN maximizes the utilization of preserved features, effective downsampling is essential to ensure high-quality inputs from the backbone; to this end, we introduce ESSamp.Information loss during downsampling operations substantially impacts the quality of shallow backbone features, which are critical for effective feature reuse in small object detection. Conventional downsampling methods, such as strided convolution, often result in severe loss of fine-grained details. Meanwhile, existing detail-preserving approaches, such as SliceSamp, suffer from computational inefficiency and limited feature expressiveness. To overcome these limitations, we introduce the Enhanced SliceSamp (ESSamp) module. ESSamp enhances feature representation through improved depthwise convolution and further optimizes the spatial rearrangement process to increase computational efficiency. The overall structure of ESSamp is illustrated in Figure~\ref{fig:essamp_structure}, and its design is described in detail below.

\begin{table*}[h!]
    \setlength{\tabcolsep}{20pt} 
    \centering
    \small
    \begin{tabular}{l|c|cc|cc}
        \toprule
        \textbf{Model} & \textbf{Size} & \textbf{AP}$_{50}$ & \textbf{AP} & \textbf{Params} & \textbf{FLOPs} \\
        \midrule
        YOLOv8-s (\citeyear{varghese2024yolov8}) & 640 & 39.6 & 23.6 & 11.2 M & 28.6 G \\
        YOLO11-s (\citeyear{khanam2024yolov11}) & 640 & 40.4 & 24.2 & 9.4 M & 21.3 G \\
        FBRT-YOLOv8-s (\citeyear{xiao2025fbrt}) & 640 & 41.7 & 25.5 & 2.9 M & 23.1 G \\
        PRNet-N(Ours) & 640 & \textbf{43.4} & \textbf{26.7} & \textbf{2.2 M} & \textbf{17.8 G} \\
        \midrule
        YOLOv8-m (\citeyear{varghese2024yolov8}) & 640 & 44.0 & 26.9 & 25.8 M & 78.4 G \\
        YOLO11-m (\citeyear{khanam2024yolov11}) & 640 & 46.1 & 28.6 & 20.1 M & 68.0 G \\
        FBRT-YOLOv8-x (\citeyear{xiao2025fbrt}) & 640 & 47.3 & 29.6 & 23.2M & 187.1 G \\
        PRNet-YOLOv8-s(Ours) & 640 & \textbf{50.4} & \textbf{31.3} & 8.2 M & 55.5 G \\
        PRNet(Ours) & 640 & 49.9 & 31.1 & \textbf{7.77 M} & \textbf{44.9 G} \\
        \midrule
        EMA attention† (\citeyear{ouyang2023efficient}) & 640 & 49.7 & 30.4 & 91.18 M & 315 G \\
        yolov9c† (\citeyear{wang2024yolov9}) & 640 & 47.6 & 29.3 & 25.3 M & 239.9 G \\
        HV-SwinViT† (\citeyear{zhao2025unified}) & 640 & 43.63 & 26.3 & 64 M & 523 G \\
        PRNet-L(Ours) & 640 & \textbf{54.1} & \textbf{34.4} & \textbf{24.6 M} & \textbf{196.8 G} \\
        \midrule
        \multicolumn{6}{l}{\textbf{Larger Input Size}} \\
        \midrule
        DQ-DETR† (\citeyear{huang2024dq}) & 800×1333 & 60.9 & 37.0 & 58 M & 904 G \\
        HV-SwinViT† (\citeyear{zhao2025unified}) & 1280 & 52.5 & 35.6 & 91.8 M & - \\
        PRNet-L(Ours) & 1024 & \textbf{61.0} & \textbf{38.3} & \textbf{24.6 M} & \textbf{505 G} \\
        \bottomrule
    \end{tabular}
    \caption{Comparison with state-of-the-art models under different resource constraints on VisDrone-Validation dataset. "--" indicates that no data were available for this item. \textbf{Bold} indicates the best results.Results marked with † are reported from the original papers, while the others are reproduced by us under the same experimental settings.}
    \label{tab:visdrone_validation}
\end{table*}
\noindent\textbf{Enhanced Feature Expression.} The primary limitation of existing detail-preserving downsampling methods lies in their insufficient feature modeling capability. Standard depthwise convolution in SliceSamp uses only a single kernel per input channel, severely limiting the ability to capture complex local patterns essential for small object discrimination. To overcome this bottleneck, we introduce enhanced depthwise convolution with depth multiplier $d$, which assigns multiple kernels to each input channel to enrich local feature representation.

This design is supported by empirical analysis and receptive field considerations. By introducing the depth multiplier $d$, the feature expressiveness is enhanced, with its impact validated through ablation studies (e.g., Table~\ref{tab:essamp_depth}). For instance, when $d=2$, the capacity for local structure modeling is effectively improved. Such an enhancement is particularly crucial for small objects, where discriminative information is extremely limited and subtle local patterns are essential for reliable detection.

\noindent\textbf{Improved Spatial Rearrangement.} In addition to the core feature enhancement, we also improve the spatial rearrangement process in SliceSamp to boost computational efficiency. SliceSamp's explicit indexing operations(e.g., $X=\text{Concat}(X_{in}[:,:,i::2,j::2]), i,j \in \{0,1\}$) incur high memory access overhead and cannot fully utilize GPU parallel computing capabilities. We replace these explicit operations with PixelUnShuffle, which improves memory coalescing and provides a constant-factor reduction in runtime while maintaining the detail-preserving property, as demonstrated in efficient sub-pixel convolutional networks \cite{shi2016real}. as shown in Equations below:
\begin{equation}
X = \text{PixelUnShuffle}(2, X_{in})
\end{equation}
\begin{equation}
Y = \text{GELU}(\text{BN}_2(W_2^{PW} * \text{GELU}(\text{BN}_1(W_1^{EDW} \odot X))))
\end{equation}

where $W_1^{EDW} \in \mathbb{R}^{4dC \times 4C \times 3 \times 3}$ is the expanded depthwise convolution kernel with groups=$4C$ and output channels of $4dC$, $d=2$; $W_2^{PW} \in \mathbb{R}^{C_{out} \times 4dC \times 1 \times 1}$ is the pointwise convolution kernel; $d$ is the depth multiplier; $*$ denotes standard convolution, $\odot$ denotes Depthwise Convolution.

\begin{figure*}[t]
\centering
\includegraphics[width=1\textwidth]{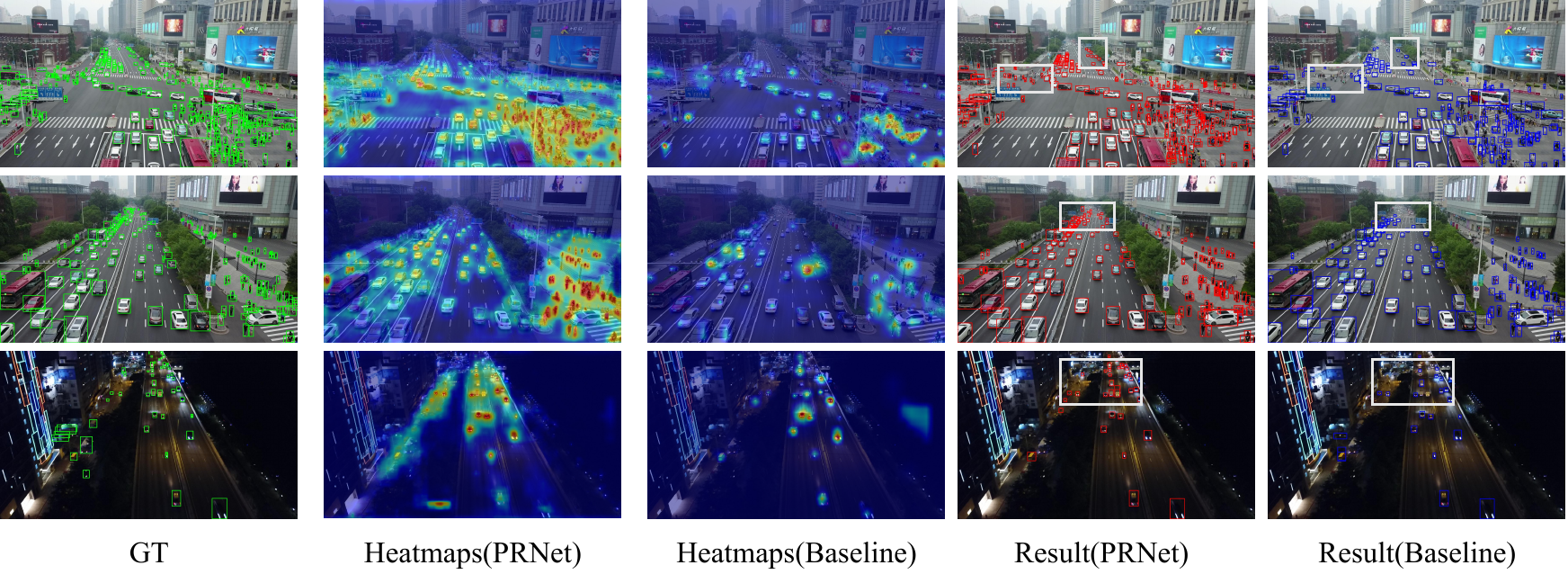} % Reduce the figure size so that it is slightly narrower than the column.
\caption{\textbf{Visualization of the detection results and heatmaps on VisDrone. } The highlighted areas represent the regions that the network is focusing on.}
\label{fig:visdrone_visualization}
\end{figure*}

Compared to traditional SliceSamp, ESSamp maintains the advantage of high-fidelity downsampling while significantly improving computational efficiency and feature expression capability, providing higher-quality feature input for PRN and achieving an optimized balance between detail preservation and computational efficiency.Channel expansion from $C$ to $4dC$ significantly enhances feature expression but also increases the computational burden of depthwise convolution. The subsequent pointwise convolution needs to compress $4dC$ back to the object channel number with a compression ratio of 4d:1, posing a risk of information bottleneck. In subsequent experiments, we conduct ablation studies on the hyperparameter $d$ to achieve an optimal balance between feature expression capability and computational overhead. This integration of ESSamp with PRN forms a comprehensive pathway for preserving and utilizing original information, enhancing overall detection performance in aerial imagery.

\section{Experiments}
\subsection{Implementation Details}
We conduct comprehensive experiments on three aerial image object detection benchmarks: VisDrone (\citeyear{du2019visdrone}), AI-TOD (\citeyear{xu2022detecting}), and UAVDT (\citeyear{du2018unmanned}). Experiments are conducted on RTX 3090 GPU. Our network is trained for 400 epochs using the stochastic gradient descent (SGD) optimizer with a momentum of 0.937, a weight decay of 0.0005, a batch size of 8, a patience of 50, and an initial learning rate of 0.01. All other configurations remain the same as in YOLO11.
\begin{table}[t]
    \centering
    \small
    \begin{tabular}{l|cc|cc}
        \toprule
        \textbf{Model} & \textbf{AP}$_{50}$ & \textbf{AP} & \textbf{Params} & \textbf{FLOPs} \\
        \midrule
        YOLOX-M† (\citeyear{ge2021yolox}) & 30.5 & 17.8 & 25.3 M & 73.75 G \\
        YOLOv7† (\citeyear{wang2023yolov7}) & 39.2 & 21.3 & 36.5 M & 103.3 G \\
        YOLOV8-M (\citeyear{varghese2024yolov8}) & 35.6 & 21.2 & 25.6 M & 78.7 G \\
        DCYOLO-M† (\citeyear{gu2025dcyolo}) & 38 & 22.3 & 33 M & 117 G \\
        \midrule
        PRNet(Ours) & \textbf{40.3} & \textbf{24.2} & \textbf{7.77 M} & \textbf{44.9 G} \\
        \bottomrule
    \end{tabular}
    \caption{Comparison of PRNet with latest methods on VisDrone test-dev dataset.}
    \label{tab:visdrone_testdev}
\end{table}

\subsection{Results on Visdrone Dataset}
\noindent\textbf{Comparison with State-of-the-art Methods.} As shown in Table~\ref{tab:visdrone_validation}, we conduct comprehensive comparisons between PRNet and existing state-of-the-art detection methods on the VisDrone validation dataset. The experimental results demonstrate that PRNet exhibits superior performance advantages across different resource constraints. For lightweight models, PRNet-N achieves 43.4\% AP$_{50}$ and 26.7\% AP with only 2.2M parameters and 17.8G FLOPs, compared to YOLO11-s, it improves detection accuracy by 3.0\% AP$_{50}$ and 2.5\% AP respectively while reducing parameters by 76.6\%. Compared to FBRT-YOLO-S with similar parameter count, AP$_{50}$ and AP are improved by 1.0\% and 0.8\% respectively. For medium-scale models, we present two variants: PRNet (based on YOLO11) achieves 49.9\% AP$_{50}$ and 31.1\% AP with 7.77M parameters and 44.9G FLOPs, while PRNet-YOLOv8-s (based on YOLOv8-s backbone) achieves even better performance of 50.4\% AP$_{50}$ and 31.3\% AP with 8.2M parameters and 55.5G FLOPs, demonstrating that even with the same YOLOv8 backbone as FBRT-YOLOv8-X, our framework achieves superior performance (50.4\% vs. 47.3\% AP$_{50}$); furthermore, as shown in Table~\ref{tab:prn_ablation_yolo}, PRN can be effectively integrated into FBRT-YOLO, further improving its detection performance. Compared to YOLO11-m, it improves detection accuracy by 3.8\% AP$_{50}$ and 2.5\% AP respectively while reducing parameters by 61.3\% and computational cost by 34.0\%. For large-scale models, PRNet-L achieves the best accuracy of 54.1\% AP$_{50}$ and 34.4\% AP, comprehensively outperforming all comparison methods, validating the detection advantages of our approach in complex aerial scenarios.

\noindent\textbf{Generalization Validation.} To further validate the generalization capability of PRNet, we conducted comparisons with the latest methods on the VisDrone test-dev dataset, with results shown in Table~\ref{tab:visdrone_testdev}. PRNet achieves 40.3\% AP$_{50}$ and 24.2\% AP, outperforming YOLOv8-M (\citeyear{varghese2024yolov8}) by 4.7\% AP$_{50}$ and 3.0\% AP, and DCYOLO-M (\citeyear{gu2025dcyolo}) by 2.3\% AP$_{50}$ and 1.9\% AP. These results validate the robust stability and superiority of PRNet in aerial image detection.
\begin{table}[t]
\centering
\small
\begin{tabular}{l|c|ccc}
\toprule
\textbf{Method} & \textbf{Size} & \textbf{AP} & \textbf{AP}$_{50}$ & \textbf{AP}$_{75}$ \\
\midrule
FFCA-YOLO† (\citeyear{10423050}) & 640 & 27.7 & 61.7 & 22.3 \\
DQ-DETR† (\citeyear{huang2024dq}) & 800×1333 & 30.2 & \textbf{68.6} & 22.3 \\
HS-FPN† (\citeyear{shi2025hs}) & 800×800 & 25.1 & 55.7 & 22.3 \\
HV-SwinVit† (\citeyear{zhao2025unified}) & 1280 & \underline{32.1} & 62.3 & \underline{29.4} \\
\midrule
PRNet(Ours) & 640 & 30.3 & 61.4 & 28.1 \\
PRNet-L(Ours) & 640 & \textbf{35.6} & \underline{67.8} & \textbf{33.1} \\
\bottomrule
\end{tabular}
\caption{Comparison of PRNet with advanced methods on AI-TOD test dataset.\textbf{Bold} indicates the best-performing method, and \underline{underline} indicates the second-best method.}
\label{tab:aitod_comparison}
\end{table}

\noindent\textbf{Qualitative Results.} To further illustrate the superior performance of PRNet in detecting small objects in aerial images, we present visualizations of heatmaps and detection results in Figure~\ref{fig:visdrone_visualization}. The heatAPs show PRNet's enhanced focus on small and densely packed objects compared to the baseline, while the detection results demonstrate more precise localization aligned with ground truth.
\begin{table}[t]
\centering
\small
\begin{tabular}{l|c|ccc}
\toprule
\textbf{Method} & \textbf{Size} & \textbf{AP} & \textbf{AP}$_{50}$ & \textbf{AP}$_{75}$ \\
\midrule
YOLO11-S (\citeyear{khanam2024yolov11})& 640 & 19.1 & 31.4 & 20.9 \\
YOLC† (\citeyear{liu2024yolc})& 1024×640 & 19.3 & 30.9 & 20.1 \\
AD-Det† (\citeyear{li2025ad})& 1024×540 & \underline{20.1} & \textbf{34.2} & \underline{21.9} \\
\midrule
PRNet(Ours)& 640 & \textbf{20.8} & \underline{32.3} & \textbf{23.8} \\
\bottomrule
\end{tabular}
\caption{Comparison of PRNet with advanced methods on UAVDT dataset.}
\label{tab:uavdt_comparison}
\end{table}
\subsection{Results on AI-TOD Dataset}
The AI-TOD dataset contains a large number of extremely small objects and high-density scenes, which impose higher demands on the fine-grained feature extraction capabilities of detection algorithms. To further validate the superiority of our method in remote sensing small object detection, we evaluated PRNet and PRNet-L on the AI-TOD test set. As shown in Table~\ref{tab:aitod_comparison}, at a compact 640×640 resolution, PRNet achieves 30.3\% AP, 61.4\% AP$_{50}$, and 28.1\% AP$_{75}$, outperforming most benchmarks. PRNet-L sets new records with 35.6\% AP and 33.1\% AP$_{75}$, surpassing DQ-DETR (\citeyear{huang2024dq}) and HV-SwinVit (\citeyear{zhao2025unified}) despite their larger input sizes. These results underscore the superior accuracy and efficiency of our framework in remote sensing scenarios.

\begin{table}[t]
    \centering
    \small
    \begin{tabular}{cc|cc|cc}
        \toprule
        \textbf{PRN} & \textbf{ESSamp} & \textbf{AP}$_{50}$ & \textbf{AP} & \textbf{Params} & \textbf{FLOPs} \\
        \midrule
        - & - & 39.0 & 23.3 & 9.4 M & 21.3 G \\
        \checkmark & - & 49.3 & 30.4 & 7.71 M & 41.1 G \\
        - & \checkmark & 40.1 & 24.1 & 9.49 M & 24.8 G \\
        \textbf{\checkmark} & \textbf{\checkmark} & \textbf{49.8} & \textbf{31.1} & \textbf{7.77 M} & \textbf{44.9 G} \\
        \bottomrule
    \end{tabular}
    \caption{Ablation study of the proposed method on VisDrone.}
    \label{tab:method_ablation}
\end{table}
\subsection{Results on UAVDT Dataset}
Table \ref{tab:uavdt_comparison} presents the comparison results on the UAVDT dataset. Our proposed method surpasses existing methods, such as YOLC(\citeyear{liu2024yolc}) and AD-Det(\citeyear{li2025ad}). Utilizing a smaller input size of 640, PRNet achieves 20.8\% AP, 32.3\% AP$_{50}$, and 23.8\% AP$_{75}$, outperforming other state-of-the-art methods in AP and AP$_{75}$ despite AD-Det's larger resolution of 1024×540. This demonstrates the effectiveness of our detection framework.

\begin{table}[t]
    \centering
    \small
    \begin{tabular}{c|cc|cc}
        \toprule
        \textbf{PRN layer} & \textbf{AP}$_{50}$ & \textbf{AP} & \textbf{Params} & \textbf{FLOPs} \\
        \midrule
        - & 39.0 & 23.3 & 9.4 M & 21.3 G \\
        0 & 45.0 & 27.6 & 7.05 M & 28.7 G \\
        1 & 49.3 & 30.4 & 7.71 M & 41.1 G \\
        2 & 51.0 & 31.8 & 8.4 M & 54.3 G \\
        3 & 51.4 & 32.2 & 9.1 M & 67.5 G \\
        \bottomrule
    \end{tabular}
    \caption{Ablation study on progressive refinement iterations.}
    \label{tab:progressive_iterations}
\end{table}

\begin{table}[t]
    \setlength{\tabcolsep}{8pt}
    \centering
    \small
    \begin{tabular}{c|c|cc|cc}
        \toprule
        \textbf{Baseline} & \textbf{Depth} & \textbf{AP}$_{50}$ & \textbf{AP} & \textbf{Params} & \textbf{FLOPs} \\
        \midrule
        PRNet & - & 49.3 & 30.4 & \textbf{7.71 M} & 41.1 G \\
        & 1 & 48.6 & 30.2 & 7.75 M & 42.9 G \\
        & 2 & \textbf{49.8} & \textbf{31.1} & 7.77 M & 44.9 G \\
        & 3 & 49.4 & 30.6 & 7.82 M & 46.9 G \\
        \bottomrule
    \end{tabular}
    \caption{Ablation study on different depths of ESSamp on VisDrone.}
    \label{tab:essamp_depth}
\end{table}

\begin{table}[t]
    \centering
    \small
    \begin{tabular}{l|c|cc|cc}
        \toprule
        \textbf{Method} & \textbf{PRN} & \textbf{AP}$_{50}$ & \textbf{AP} & \textbf{Params} & \textbf{FLOPs} \\
        \midrule
        YOLOv5s & - & 40.3 & 23.9 & 9.1 M & \textbf{23.8 G} \\
        & \checkmark & \textbf{47.4} & \textbf{29.2} & \textbf{6.7 M} & 42.7 G \\
        YOLOv5m & - & 44.4 & 27.1 & 25.9 M & 78.9 G \\
        \midrule
        YOLOv8s & - & 40.5 & 24.4 & 11.1 M & \textbf{28.7 G} \\
        & \checkmark & \textbf{48.3} & \textbf{30.2} & \textbf{8.3 M} & 54.1 G \\
        YOLOv8m & - & 44.0 & 26.9 & 25.8 M & 78.4 G \\
        \midrule
        YOLO11s & - & 39.0 & 23.3 & 9.4 M & \textbf{21.3 G} \\
        & \checkmark & \textbf{49.3} & \textbf{30.4} & \textbf{7.71 M} & 41.1 G \\
        YOLO11m & - & 46.1 & 28.6 & 20.1 M & 68.0 G \\
        \midrule
        FBRT-YOLOv8-s & - & 41.7 & 25.5 & 2.9 M & \textbf{23.1 G} \\
        & \checkmark & \textbf{47.7} & \textbf{29.2} & \textbf{2.2 M} & 40.4 G \\
        FBRT-YOLOv8-m & - & 45.9 & 28.4 & 7.2 M & 58.7 G \\
        \midrule
        RT-DETR-R50 & - & 28.9 & 16.1 & 42 M & \textbf{136 G} \\
        & \checkmark & \textbf{32.1} & \textbf{18.3} & \textbf{39.9 M} & 176.4 G \\
        RT-DETR-R101 & - & 31.5 & 17.8 & 60.9 M & 186.3 G \\
        \bottomrule
    \end{tabular}
    \caption{Ablation study of PRN on different YOLO detectors on VisDrone. "-" indicates PRN method is not used.}
    \label{tab:prn_ablation_yolo}
\end{table}

\subsection{Ablation experiments}
We conduct ablation experiments on the VisDrone dataset using YOLO11-S as the baseline to validate PRNet’s core components.

\noindent\textbf{Effect of Key Components.} As shown in Table~\ref{tab:method_ablation}, the baseline YOLO11s achieves 39.0\% AP$_{50}$ and 23.3\% AP. Adding PRN alone improves AP$_{50}$ by 10.3\% to 49.3\% and AP by 7.1\% to 30.4\%, while reducing parameters by 18\% (9.4M to 7.71M). Using ESSamp alone yields modest gains (40.1\% AP$_{50}$, 24.1\% AP). Combining PRN and ESSamp achieves the best performance (49.8\% AP$_{50}$, 31.1\% AP), with 7.77M parameters and 44.9G FLOPs, demonstrating their synergistic effect. Although PRNet increases FLOPs by 110.7\% (21.3G to 44.9G) compared to the baseline, this computational overhead is strategically justified: PRN's iterative refinement operates primarily on high-resolution features where small objects reside, directly translating increased computation into substantial accuracy gains (10.8\% AP$_{50}$, 7.8\% AP), while our overall framework maintains superior efficiency compared to state-of-the-art methods—for instance, achieving comparable accuracy to YOLO11-m (46.1\% AP$_{50}$) while using 34.0\% fewer FLOPs (44.9G vs. 68.0G), demonstrating an advantageous accuracy-efficiency trade-off for aerial small object detection.

\noindent\textbf{Effect of Progressive Refinement Stages.} Table~\ref{tab:progressive_iterations} shows that increasing PRN iterations from 0 to 3 improves AP$_{50}$ from 45.0\% to 51.4\% and AP from 27.6\% to 32.2\%. Although further increasing the repetition count can continue to improve accuracy, the computational overhead also grows significantly. Therefore, considering real-time performance, we select 1 repetition as the optimal configuration.

\noindent\textbf{Effect of Depth Multiplier in ESSamp.} Table~\ref{tab:essamp_depth} evaluates ESSamp’s depth multiplier. Depth=2 achieves the best performance (49.8\% AP$_{50}$, 31.1\% AP), improving over depth=1 (equivalent to SliceSamp) by 1.2\% AP$_{50}$ and 0.9\% AP. Further increasing depth to 3, although theoretically enhancing feature expression further, causes the dramatic increase in channel numbers to require larger compression ratios in subsequent pointwise convolutions, leading to key information loss and reduced detection accuracy. Therefore, we select depth=2 as the optimal configuration for ESSamp.

\noindent\textbf{Effect of PRN Generalizability.} Table~\ref{tab:prn_ablation_yolo} validates PRN’s versatility across YOLOv5s, YOLOv8s, YOLO11s, FBRT-YOLOv8-s and RT-DETR-R50. After introducing PRN to all tested detectors, detection accuracy is significantly improved. YOLO11s+PRN achieves the highest gains, improving AP$_{50}$ by 10.3\% and AP by 7.1\% while reducing parameters by 18\%. Notably, when applying PRN to the state-of-the-art FBRT-YOLOv8-s baseline (41.7\% AP$_{50}$), our method achieves 47.7\% AP$_{50}$ and 29.2\% AP—a substantial improvement of 6.0\% AP$_{50}$ and 3.7\% AP—while reducing parameters from 2.9M to 2.2M, demonstrating that PRN can further enhance already optimized architectures. Furthermore, even compared to larger m-series models, lightweight models with PRN can achieve superior detection performance under fewer resource constraints, validating the universality and effectiveness of PRN across different architectures.

\noindent Due to page constraints, additional experimental results and visualizations are provided in the \textbf{Appendix}.

\section{Conclusion}
In this paper, we address the challenge of information loss in small object detection for aerial imagery by proposing PRNet, a novel real-time detection framework that prioritizes the preservation and efficient utilization of original shallow spatial features. The framework introduces two key innovations: the Progressive Refinement Neck (PRN), which enables multi-stage backbone feature reuse and iterative refinement for enhanced spatial-semantic alignment, and the Enhanced SliceSamp (ESSamp), which optimizes downsampling through improved spatial rearrangement and augmented depthwise convolution to minimize detail degradation. Extensive experiments on the VisDrone, AI-TOD, and UAVDT datasets demonstrate that PRNet achieves superior detection accuracy while maintaining competitive computational efficiency, outperforming state-of-the-art methods across various resource constraints.

\bibliography{aaai2026}

\end{document}